\documentclass[9pt]{IEEEtran}
\IEEEoverridecommandlockouts
\usepackage{cite}
\usepackage{amsmath,amssymb,amsfonts}
\usepackage{graphicx}
\usepackage{textcomp}
\usepackage{amsmath}
\usepackage{booktabs} 
\usepackage{multirow}
\usepackage{hyperref}       
\usepackage{url}            
\usepackage{booktabs}       
\usepackage{amsfonts}       
\usepackage{nicefrac}       
\usepackage{microtype}      
\usepackage{xcolor}         
\usepackage{amsmath}
\usepackage{tikz}

\usepackage{orcidlink}
\usepackage{comment}
\usepackage{stfloats}
\usepackage{setspace}
\usepackage{amssymb}
\usepackage{braket}
\usepackage{xcolor}
\usepackage[compact]{titlesec}
\usepackage[font=footnotesize,skip=0pt]{caption}

\titlespacing\section{0pt}{0.3\baselineskip}{0.2\baselineskip}
\titlespacing\subsection{0pt}{0.2\baselineskip}{0.1\baselineskip}
\titlespacing\subsubsection{0pt}{0.15\baselineskip}{0.1\baselineskip}

\usepackage{algpseudocode} 
\usepackage[most]{tcolorbox}
\tcbuselibrary{skins, breakable, hooks}

\newtcolorbox{algobox}[1][]{
  enhanced, breakable,
  colback=white, colframe=black,
  boxrule=0.6pt, arc=2pt,
  left=8pt, right=8pt, top=8pt, bottom=8pt,
  attach boxed title to top left={yshift=-2pt, xshift=6pt},
  boxed title style={
    colback=black!5,   
    colframe=black,
    boxrule=0.6pt,
    arc=2pt
  },
  coltitle=black,       
  title={\textbf{#1}}
}
\usepackage{enumitem}
\setlist[itemize]{leftmargin=*}%
\setlist[enumerate]{leftmargin=*}%

\def\BibTeX{{\rm B\kern-.05em{\sc i\kern-.025em b}\kern-.08em
    T\kern-.1667em\lower.7ex\hbox{E}\kern-.125emX}}

\usepackage[compact]{titlesec}

\titlespacing\section{0pt}{0.3\baselineskip}{0.2\baselineskip}
\titlespacing\subsection{0pt}{0.2\baselineskip}{0.1\baselineskip}
\titlespacing\subsubsection{0pt}{0.15\baselineskip}{0.1\baselineskip}

\usepackage{fancyhdr}
\fancypagestyle{firstpage}{
   \fancyhf{} 
   \fancyhead[C]{To appear in 2026 Design, Automation and Test in Europe (DATE) Conference Proceedings} 
}

\begin{document}

\title{FAQNAS: FLOPs-aware Hybrid Quantum Neural Architecture Search using Genetic Algorithm\\
\vspace{-5pt}
}


\author{\IEEEauthorblockN{Muhammad Kashif \IEEEauthorrefmark{1}\IEEEauthorrefmark{2}, Shaf Khalid\IEEEauthorrefmark{1}\IEEEauthorrefmark{2}, Alberto Marchisio\IEEEauthorrefmark{1}\IEEEauthorrefmark{2}, Nouhaila Innan\IEEEauthorrefmark{1}\IEEEauthorrefmark{2},
Muhammad Shafique\IEEEauthorrefmark{1}\IEEEauthorrefmark{2}}

\IEEEauthorblockA{\IEEEauthorrefmark{1} \normalsize eBrain Lab, Division of Engineering, New York University Abu Dhabi, PO Box 129188, Abu Dhabi, UAE\\}
\IEEEauthorblockA{\IEEEauthorrefmark{2} \normalsize Center for Quantum and Topological Systems, NYUAD Research
Institute, New York University Abu Dhabi, UAE}

Emails: \{muhammadkashif, sk10741, alberto.marchisio, nouhaila.innan, muhammad.shafique\}@nyu.edu

\vspace{-30pt}
}

\maketitle

\thispagestyle{firstpage}
\begin{abstract}
Hybrid Quantum Neural Networks (HQNNs), which combine parameterized quantum circuits with classical neural layers, are emerging as promising models in the noisy intermediate-scale quantum (NISQ) era. While quantum circuits are not naturally measured in floating point operations (FLOPs), most HQNNs (in NISQ era) are still trained on classical simulators where FLOPs directly dictate runtime and scalability. Hence, FLOPs represent a practical and viable metric to measure the computational complexity of HQNNs. 
In this work, we introduce FAQNAS, a FLOPs-aware neural architecture search (NAS) framework that formulates HQNN design as a multi-objective optimization problem balancing accuracy and FLOPs. Unlike traditional approaches, FAQNAS explicitly incorporates FLOPs into the optimization objective, enabling the discovery of architectures that achieve strong performance while minimizing computational cost. Experiments on five benchmark datasets (MNIST, Digits, Wine, Breast Cancer, and Iris) show that quantum FLOPs dominate accuracy improvements, while classical FLOPs remain largely fixed. Pareto-optimal solutions reveal that competitive accuracy can often be achieved with significantly reduced computational cost compared to FLOPs-agnostic baselines. Our results establish FLOPs-awareness as a practical criterion for HQNN design in the NISQ era and as a scalable principle for future HQNN systems.\\
\vspace{-13pt}
\end{abstract}

\begin{IEEEkeywords}
Quantum Machine Learning, Hybrid Quantum Neural Networks, FLOPs, Quantum Neural Architecture Search\\
\vspace{-15pt}
\end{IEEEkeywords}

\begin{spacing}{1.0}
\section{Introduction} \label{sec:intro}
\begin{figure}
    \centering
    \vspace{-4pt}
    \includegraphics[width=0.9\linewidth]{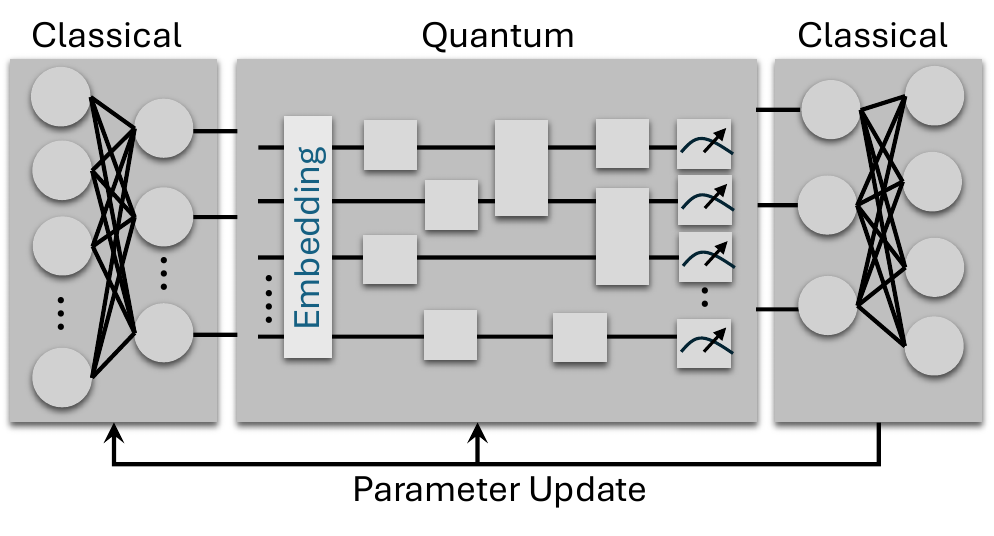}
    \vspace{-0.3cm}
    \caption{An illustration of a Hybrid Quantum Neural Network architecture. Typically Parametrized Quantum Circuits are sandwiched between classical neuron layers with classical optimization loop.}
    \label{fig:HQNN}
    \vspace{-5pt}
\end{figure}

The rapid advancement of quantum computing has spurred significant interest in Quantum Machine Learning (QML) as a paradigm that may offer computational advantages over classical methods in the near term \cite{Schuld_2014,kashif_position}. Hybrid Quantum Neural Networks (HQNNs), which integrate parameterized quantum circuits (PQCs) with classical neural layers (Fig. \ref{fig:HQNN}), have emerged as promising candidates for leveraging noisy intermediate-scale quantum (NISQ) devices \cite{kashif:2021_DSE}. These architectures combine the representational power of quantum states with the scalability of classical models, making them attractive for practical learning tasks \cite{kashif_demonstrating}. However, the design of effective HQNN architectures remains a non-trivial challenge. Unlike classical deep learning, where architecture search has been extensively studied \cite{salmani2025systematic,salehin2024automl}, architecture design in HQNNs is still relatively underexplored. Existing efforts in quantum architecture search have primarily targeted performance optimization, either by customizing circuit designs for specific devices or by developing strategies to mitigate hardware noise and scalability constraints, with the aim of achieving higher accuracy \cite{du2022quantum,li2023eqnas,li2024balanced,kashif2024nrqnn,kashif2024hqnet, dutta2025qas}. As most of these approaches are performance-oriented, our methodology fills this gap by giving equal importance to predictive accuracy and computational resource efficiency, thereby enabling a more balanced design.   

As HQNNs continue to evolve, a central question emerges: how can we design architectures that deliver competitive performance while keeping computational demands manageable? In practice, model selection for HQNNs involves conflicting objectives. On one hand, increasing the number of trainable parameters or circuit depth can improve expressivity and classification performance\cite{Kashif:2023_unified}. On the other hand, deeper circuits and larger parameter spaces are costly as they exacerbate barren plateaus in the optimization landscape and introduce significant computational overhead\cite{McClean:2018,kashif:2023impact, kashif:2024resqunns,Kashif_resqnets_2024, kashif_alleviating}. This overhead is particularly critical in the NISQ era, where most HQNNs are trained and benchmarked on classical simulators. In such settings, floating-point operations (FLOPs) directly determine runtime feasibility, memory footprint, and scalability\cite{kashif:2025computational}. Even when real quantum hardware is used for circuit execution, the hybrid training loop, including gradient estimation, optimization, and loss evaluation, remains largely classical and FLOPs-intensive. Thus, FLOPs provide a practical and unifying measure of computational cost in HQNNs.

Beyond their current role in simulation, FLOPs-aware principles also have forward-looking relevance. As hybrid workloads increasingly integrate quantum processors with large-scale classical infrastructures, the classical component will remain an indispensable bottleneck, from data encoding to iterative optimization. FLOPs-aware design therefore serves as a hardware-agnostic efficiency metric, guiding architectures that are not only viable for present-day simulators but also scalable for future hybrid quantum–classical deployments.
Motivated by these observations, this work investigates the FLOPs-aware search of HQNN architectures under two objectives: minimizing FLOPs and maximizing accuracy. By framing HQNN design as a multi-objective optimization problem, we use genetic algorithm \cite{forrest1996genetic}, and aim to uncover trade-offs that illuminate the balance between efficiency and performance in hybrid models. To demonstrate generality, we evaluate our approach on five benchmark datasets of varying complexity: Iris and Wine (structured, small-scale datasets), Breast Cancer and Digits (medium-scale datasets), and MNIST (a canonical image classification task)\footnote{Although relatively small in scale, these datasets are widely adopted in the QML community as standard benchmarks. The use of larger datasets is generally infeasible due to the prohibitive computational overhead associated with simulating quantum circuits, where the cost grows exponentially with the number of qubits and training iterations.}. These datasets allow us to examine how optimal HQNN architectures adapt across problem domains and scales, providing insights into resource-efficient design for both simulators and emerging hardware.


\subsection{Our Contributions}

Our contributions are summarized as follows:
\begin{itemize}
    
\item \textbf{Motivation for FLOPs-aware HQNNs:} We argue for the relevance of FLOPs-aware design in HQNNs, showing how FLOPs serve as a practical metric for simulator-based training in the NISQ era and as a guiding principle for scalable quantum–classical integration.

\item \textbf{FLOPs-aware NAS framework:} We develop a multi-objective neural architecture search (NAS) framework that explicitly incorporates FLOPs into the optimization process, balancing predictive accuracy with computational efficiency for HQNNs.

\item \textbf{Dataset-driven evaluation:} We apply our FLOPs-aware HQNN search to five benchmarks (MNIST, Digits, Wine, Breast Cancer, and Iris), demonstrating how FLOPs-aware optimization adapts to dataset complexity and identifies Pareto-optimal architectures.

\item \textbf{Insights into FLOPs–accuracy trade-offs:} Our analysis highlights that classical FLOPs remain mostly fixed by design, while quantum FLOPs drive accuracy improvements, and shows how FLOPs-awareness helps avoid over-provisioned architectures.

\item \textbf{Blueprint for resource-efficient HQNNs:} We provide a principled approach for designing hybrid networks that balance accuracy and efficiency, offering practical benefits for simulation-based exploration today and scalable deployment strategies for future quantum–classical computing.

\end{itemize}

By addressing both performance and resource considerations, our study sheds light on the practical design space of HQNNs. We anticipate that this work will contribute to the growing discourse on making QML models not only accurate but also hardware-efficient and trainable in realistic settings.



\section{Our Methodology}

\begin{figure*}
    \centering
    \includegraphics[width=0.98\linewidth]{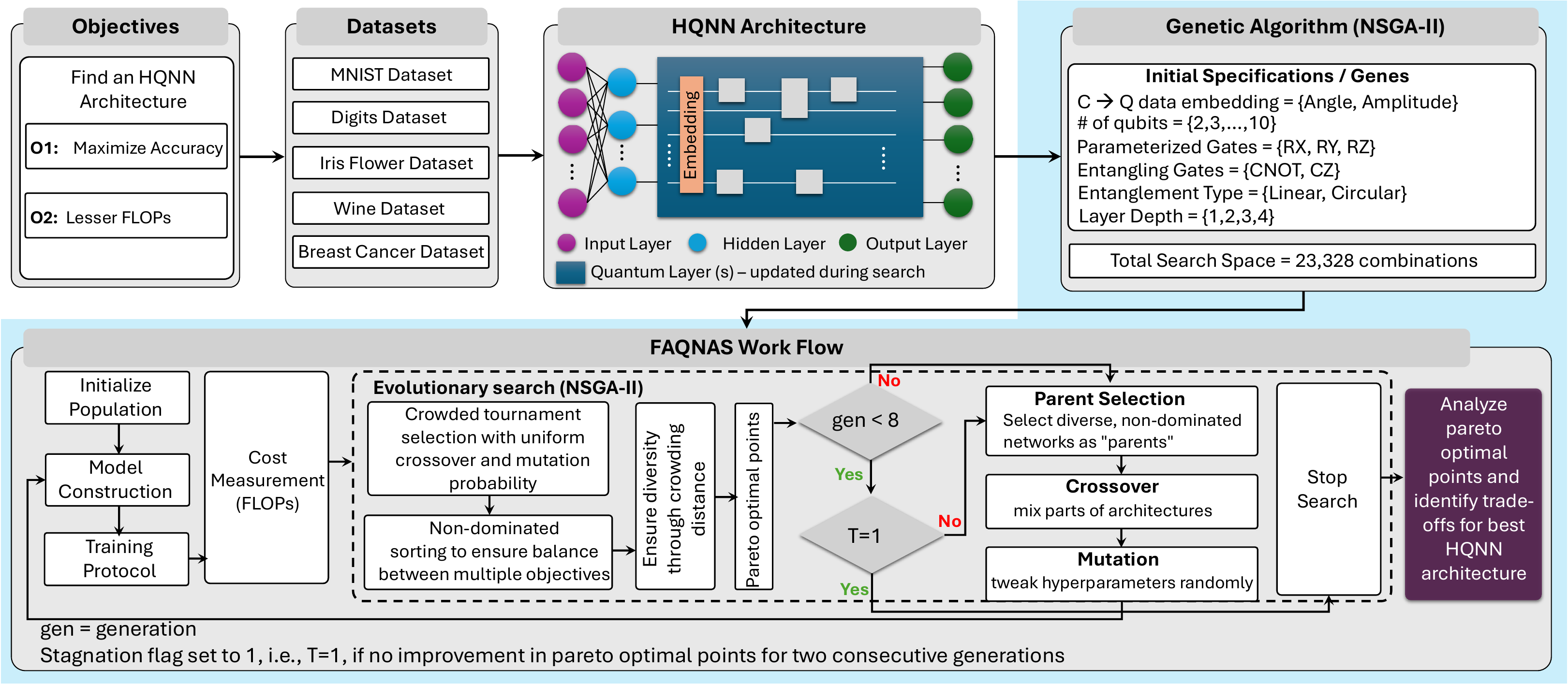}
    \vspace{-1pt}
    \caption{Our Methodology for FLOPs-aware Hybrid Quantum Neural Network (HQNN) architecture search. Candidate HQNNs are evaluated on accuracy and FLOPs across multiple datasets. The search evolves quantum layer configurations (encoding, qubits, gates, entanglement, depth) from a space of $23,328$ combinations using non-dominated sorting, crossover, and mutation. Pareto-optimal architectures are identified to balance performance and efficiency.}
    \label{fig:methodology_flopNas}
\end{figure*}

\subsection{Objective}
The primary objective of the proposed framework is to systematically explore and identify HQNN architectures that achieve a favorable balance between accuracy and computational efficiency. Specifically, the framework aims to maximize model performance while simultaneously minimizing resource consumption in terms of FLOPs. By incorporating Pareto-based multi-objective optimization, the methodology ensures that the selected architectures are not only accurate but also scalable and practical for deployment in resource-constrained environments.


\subsection{Datasets}
To evaluate the effectiveness and generalizability of the proposed framework, we employed five widely used benchmark datasets that vary in complexity and dimensionality:

\subsubsection{MNIST Handwritten Digits Dataset}
The MNIST dataset \cite{lecun2010mnist}is a large-scale benchmark consisting of $70,000$ grayscale images of handwritten digits ($0–9$), each with a resolution of 
$28\times 28$ pixels. It includes $10$ output classes, corresponding to the ten digit categories. The models are trained on $60,000$ images and tested on $10,000$ images. It serves as a standard dataset for image classification tasks and provides a relatively high-dimensional feature space, which is suitable for testing the scalability of HQNN architectures.

\subsubsection{Digits Dataset}
The Digits dataset \cite{pedregosa2011scikit} is a medium-scale benchmark comprising $1,797$ grayscale images of handwritten digits ($0–9$). Each image has a resolution of $8\times 8$ pixels, yielding a significantly lower-dimensional feature space compared to MNIST’s $28\times 28$ resolution. The dataset includes $10$ output classes, corresponding to the digit categories. Unlike MNIST, which contains $70,000$ samples and serves as a high-dimensional benchmark for large-scale image classification, the Digits dataset is relatively small in size and dimensionality. This makes it particularly useful for rapid prototyping and low-resource experimentation with HQNN architectures, while MNIST is better suited for evaluating scalability under more complex input spaces.

\subsubsection{Iris Flower Dataset}
The Iris dataset \cite{pedregosa2011scikit} is a classical benchmark in ML comprising $150$ samples from three species of Iris flowers (setosa, versicolor, and virginica). Each sample is represented by four numerical features: sepal length, sepal width, petal length, and petal width. Its small size ($3$ classes) and low dimensionality make it particularly useful for testing the behavior of HQNNs in constrained feature spaces.

\subsubsection{Wine Dataset}
The Wine dataset \cite{pedregosa2011scikit} contains $178$ samples of three wine cultivars. Each sample is described by $13$ continuous physicochemical features, such as alcohol content, flavanoids, and color intensity. This dataset includes $3$ classes, corresponding to the wine cultivars, and represent a medium-sized classification challenge and enables assessment of the framework on structured tabular data.

\subsubsection{Breast Cancer Dataset}
The Breast Cancer Wisconsin Diagnostic dataset \cite{pedregosa2011scikit} consists of $569$ samples describing cell nuclei features computed from digitized images of breast tissue. Each sample is characterized by $30$ real-valued features, which capture properties such as radius, texture, perimeter, area, and smoothness of the cell nuclei. The dataset is divided into two output classes: malignant and benign tumors. It is a well-established benchmark for binary classification tasks in biomedical ML. The relatively small number of samples, combined with a moderate feature dimension, makes it a useful dataset for testing the generalization ability of HQNNs under constrained data availability.

Collectively, these datasets cover a diverse range of modalities, images, low-dimensional tabular data, and medium-dimensional physicochemical data, thereby providing a comprehensive testbed for evaluating the trade-offs between accuracy and resource efficiency in HQNN architectures.


\subsection{HQNN Architecture} \label{subsec:HQNN_architecture}
The HQNN architecture that we consider has three components (as shown in Fig. \ref{fig:HQNN}), details of which are presented below:

\begin{enumerate}
    \item \textbf{Classical Pre-processing:} The classical front-end consists of a fully connected (linear) layer that prepares raw input data for quantum processing. Given an input sample represented as a feature vector of dimensionality $d$, this layer projects it onto an $n_{\text{qubits}}$-dimensional feature space through multiplication with a weight matrix of size $d\times n_{\text{qubits}}$. The resulting vector of length $n_{\text{qubits}}$ serves as the embedding that is fed into the quantum layer. This linear transformation serves a dual purpose: dimensionality reduction for high-dimensional inputs and feature expansion for low-dimensional inputs, thereby aligning the data representation with the quantum circuit size.

    \item \textbf{Quantum Layers:} The quantum layer forms the core of the hybrid architecture and is implemented using PennyLane as a QNode wrapped as a PyTorch layer\footnote{\url{https://pennylane.ai/qml/demos/tutorial\_qnn\_module\_torch}}, enabling seamless integration with classical optimization routines. The input vector from the front-end is embedded into the quantum state using one of two encoding strategies: angle embedding or amplitude embedding (Eq. (\ref{AngleE_eq1}) and Eq. (\ref{AmpE_eq1}), respectively), as selected by the Genetic Algorithm (GA). The PQC then applies up to a configurable number of layers of parameterized single-qubit rotations. For each qubit and each layer, the GA selects the gate type from the set $\{R_X(\theta),R_Y(\theta),R_Z(\theta)\}$. The rotation angles are trainable parameters organized in a tensor of shape $(\text{max\_layers},n_{\text{qubits}})$. 
    \begin{equation}
    \ket{x} = \bigotimes_{i=1}^{N} cos(x_i)\ket{0}+sin(x_i)\ket{1}
    \label{AngleE_eq1}
\end{equation}
\begin{equation}
      \Bigg(\begin{smallmatrix}
x_1\\
\vdots \\
x_{2^n}\\ 
\end{smallmatrix}\Bigg)  
\leftrightarrow \ket{\psi_x} = \sum_{j=1}^{2^n} x_j \ket{j}  
\label{AmpE_eq1}
\end{equation}
    Entanglement is introduced to capture correlations across qubits. Depending on the GA configuration, one or more rounds of entangling operations are applied in either a linear or circular topology (Fig. \ref{fig:entanglement_topo}), using two-qubit gates chosen from $\{CNOT,CZ\}$.
    
    Finally, measurement is performed in the computational basis: the expectation value of the Pauli-Z operator is obtained for each qubit, resulting in an $n_{\text{qubits}}$-dimensional feature vector. This vector represents the quantum-transformed embedding and is passed to the classical post-processing stage.

    \begin{figure}
        \centering
        \includegraphics[width=0.8\linewidth]{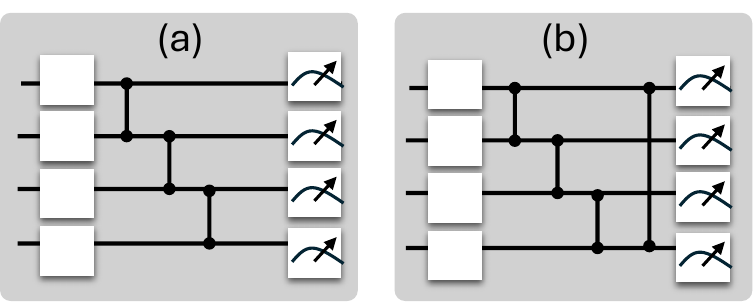}
        \caption{An Illustration of Entanglement Topology used in this paper. (a) Linear Topology (b) Circular Topology.}
        \label{fig:entanglement_topo}
    \end{figure}
    
    \item \textbf{Classical Post-Processing:} The classical post-processing module is responsible for mapping the quantum-transformed feature vector to the final prediction space. A fully connected layer projects the $n_{\text{qubits}}$-dimensional output into a vector of logits corresponding to the number of classes in the target dataset. A log-softmax activation is applied to convert the logits into log-probabilities, which are then used for training and evaluation through a negative log-likelihood (NLL) loss.
\end{enumerate}


\subsection{Genetic Algorithm}
Genetic algorithms (GAs) are powerful evolutionary optimization techniques inspired by natural selection and survival of the fittest. Among them, the Non-dominated Sorting Genetic Algorithm II (NSGA-II) has emerged as one of the most widely adopted approaches for solving multi-objective optimization problems \cite{ma2023_GA_review}. NSGA-II efficiently balances convergence toward the Pareto-optimal front with diversity preservation using mechanisms such as fast non-dominated sorting, elitism, and crowding distance. Its computational efficiency and robustness make it particularly suitable for complex, real-world applications where trade-offs among conflicting objectives must be carefully managed. By maintaining a diverse set of high-quality solutions, NSGA-II provides researchers and practitioners with deeper insights into the optimization landscape.

\begin{table}[h!]
\centering
\caption{Search space specifications for genetic algorithm encoding}
\label{tab:GA_specs}
\begin{tabular}{|p{3.8cm}|p{3.8cm}|}
\hline
\textbf{Parameter} & \textbf{Values} \\ \hline
Number of Qubits ($n_{\text{qubits}}$) & \{2, 3, 4, 5, 6, 7, 8, 9, 10\} \\ \hline
Embedding Method & \{Angle, Amplitude\} \\ \hline
Parameterized Rotation Gates & \{$R_X(\theta), R_Y(\theta), R_Z(\theta)$\} \\ \hline
Entangling Gate & \{CNOT, CZ\} \\ \hline
Entanglement Topology & \{Linear, Circular\} \\ \hline

Number of  Layers & \{1, 2, 3, 4\} \\ \hline
\end{tabular}
\end{table}
\subsubsection{Initial Specifications}
In the proposed FAQNAS framework, each individual in the population is encoded as a chromosome where the genes correspond to quantum circuit hyperparameters. The search space is defined over several quantum-specific parameters that form the genetic material. The number of qubits is varied from 2 to 10, enabling exploration of scalability across circuit widths. For classical to quantum data embedding, two schemes are considered: angle embedding and amplitude embedding, which encode classical data into quantum states with different resource requirements. 

Entangling operations are introduced by both gate type, either CNOT or CZ, and entanglement topology, chosen from linear or circular connectivity. The number of entangling layers is allowed to vary from $1$ to $4$, thereby controlling circuit depth and entanglement capacity. Additionally, parameterized rotation gates are randomly drawn from the set $\{R_X(\theta),R_Y(\theta),R_Z(\theta)\}$. Together, these parameters form the genes of the evolutionary search, where different combinations represent candidate architectures subject to optimization. The combination of these parameters yields a total search space of $23,328$ unique quantum circuit configurations. This structured representation ensures that the genetic algorithm can systematically evolve candidate HQNN architectures by recombining and mutating different parameter sets, while NSGA-II guides the search toward Pareto-optimal solutions with respect to accuracy and FLOPs. A summary of the specifications that govern the GA search space is presented in Table \ref{tab:GA_specs}.


\subsubsection{FAQNAS Workflow}
The evolutionary process for identifying optimal HQNN architectures follows the Non-dominated Sorting Genetic Algorithm II (NSGA-II), the complete pseudocode of FAQNAS is presented below followed by a step-by-step explanation: 

\begin{algobox}[FLOPs-Aware NSGA-II for HQNN Architecture Search]
\begin{algorithmic}[1]
\label{algo:NSGA}
\footnotesize
\State \textbf{Inputs:} pop size $P$, generations $G$, $p_c$, $p_m$, stagnation $T$, search space $\mathcal{S}$

\State \textbf{Objectives:} $f(\mathbf{c})=(F^{\mathrm{Q}},\,1-\mathrm{Acc}_{\mathrm{val}},\,\mathrm{Params})$, with $F^{\mathrm{Q}}=\max\{F^{\mathrm{Total}}-F^{\mathrm{Classical}},0\}$
\State $\mathcal{P}\gets \textproc{InitializePopulation}(P,\mathcal{S})$
\ForAll{$\mathbf{c}\in\mathcal{P}$} \State $\mathbf{c}.f\gets \textproc{Evaluate}(\mathbf{c})$ \EndFor
\State $\mathcal{F}\gets \textproc{FastNonDominatedSort}(\mathcal{P})$;\; \textproc{ComputeCrowdingDistance}$(F),~\forall F\in\mathcal{F}$
\State $\mathcal{F}^*_{\text{prev}}\gets$ rank-0 objective tuples;\; $\text{stagn}\gets 0$
\For{$g=1$ \textbf{to} $G$}
  \State $\mathcal{O}\gets\emptyset$
  \For{$i=1$ \textbf{to} $P$}
    \State $p_1\gets \textproc{CrowdedTournamentSelect}(\mathcal{P})$;\; $p_2\gets \textproc{CrowdedTournamentSelect}(\mathcal{P})$
    \State $\mathbf{c}\gets \textproc{UniformCrossover}(p_1.\mathbf{c},p_2.\mathbf{c})$ \textbf{w.p.} $p_c$ \textbf{else} $p_1.\mathbf{c}$
    \State $\mathbf{c}\gets \textproc{PointwiseMutate}(\mathbf{c},p_m,\mathcal{S})$
    \State $\mathbf{c}.f\gets \textproc{Evaluate}(\mathbf{c})$;\; $\mathcal{O}\gets \mathcal{O}\cup\{\mathbf{c}\}$
  \EndFor
  \State $\mathcal{U}\gets \mathcal{P}\cup\mathcal{O}$;\; $\mathcal{F}\gets \textproc{FastNonDominatedSort}(\mathcal{U})$
  \State $\mathcal{P}_{\text{new}}\gets\emptyset$
  \ForAll{$F\in\mathcal{F}$}
    \State \textproc{ComputeCrowdingDistance}$(F)$
    \If{$|\mathcal{P}_{\text{new}}|+|F|\le P$} \State $\mathcal{P}_{\text{new}}\gets\mathcal{P}_{\text{new}}\cup F$
    \Else
      \State \textproc{SortByCrowdingDistanceDesc}$(F)$;\; take first $P-|\mathcal{P}_{\text{new}}|$; \textbf{break}
    \EndIf
  \EndFor
  \State $\mathcal{P}\gets \mathcal{P}_{\text{new}}$
  \State $\mathcal{F}^*\gets$ rank-0 tuples of $\mathcal{P}$;\;
         $\text{stagn}\gets \text{stagn}+1$ \textbf{if} $\mathcal{F}^*=\mathcal{F}^*_{\text{prev}}$ \textbf{else} $0$;\;
         $\mathcal{F}^*_{\text{prev}}\gets \mathcal{F}^*$
  \If{$\text{stagn}\ge T$} \State \textbf{break} \EndIf
\EndFor
\State \textbf{Return:} rank-0 Pareto set; Top-$K$; optional single compromise (min $\ell_2$ after max-norm)
\end{algorithmic}
\end{algobox}

    \paragraph{Initialize Population} To start the search, an initial population is created, where each genome $\mathrm{\textbf{c}}$ independently samples each gene from Table \ref{tab:GA_specs} to produce a diverse starting pool. As the search progresses, all the candidate architectures are defined by a unique set of hyperparameters (embedding type, number of qubits, gate choices, entanglement structure, and layer depth) sampled from the search space.

    \paragraph{Model Construction} For a given $\mathbf{c}$, we build a three-stage HQNN similar to that shown in Fig. \ref{fig:HQNN} and explained in Section \ref{subsec:HQNN_architecture}: (i) a classical linear (pre-processing) layer that flattens the input (e.g., $28{\times}28$ for MNIST) and projects to $\mathbb{R}^{n_{\text{qubits}}}$, (ii) a quantum module with the chosen feature map (Angle Embedding or Amplitude Embedding; for the latter we zero-pad and normalize to length $2^{n_{\text{qubits}}}$), $L$ layers of per-wire single-qubit rotations with the specified gate types, and $n_{\text{ent}}$ entangling layers under the selected topology; measurements are $Z$-expectations per qubit, and (iii) a linear classical (postprocessing) layer mapping $\mathbb{R}^{n_{\text{qubits}}}\to \mathbb{R}^{n}$ (where $n$ is number of classes in the dataset) followed by a negative log-likelihood (NLL) loss.

    \paragraph{Training Protocol for Candidate Architectures} Each architecture in the population is trained on the selected datasets to evaluate its learning ability and generalization performance. We use Adam with a fixed learning rate (e.g., $10^{-3}$), batch size $64$, and train each candidate $5$ epochs. We report the best validation accuracy ($\text{Acc}_{\text{val}}$) over epochs, to reduce variance from early-epoch noise. Random seeds are fixed for reproducibility.

    \paragraph{Cost measurement (FLOPs)} To make FLOPs comparable across candidates, we measure \emph{one-sample} forward+backward cost with \texttt{PyTorch profiler\footnote{\url{https://docs.pytorch.org/docs/stable/profiler.html}}}. For each genome, (i) We run the full hybrid model to obtain total number FLOPs ($F^{\text{Total}}$) and (ii) run a classical-only baseline (pre and post-processing classical layers) to obtain the FLOPs consumption of classical layers $F^{\text{Classical}}$. The quantum contribution is $F^{\text{Q}}=\max\{F^{\text{Total}}-F^{\text{Classical}},0\}$. This isolates the incremental cost of the quantum layer while holding classical work constant except for the width of hidden layer coupling to $n_{\text{qubits}}$.

    \paragraph{Multi-objective formulation} The trained networks are then assessed with respect to the optimization objectives: maximizing classification accuracy and minimizing FLOPs. We minimize the vector $(F^{\text{Q}},\,1-\text{Acc}_{\text{val}},\,\text{Params})$. The first term embodies the FLOPs-aware design principle and the second preserves predictive performance.
    
    \paragraph{Evolutionary search (NSGA-II)} 
    \begin{itemize}
        \item We initialize $P$ genomes by uniform sampling of search space $\mathcal{S}$ and evaluate their objectives.
        
        \item Each generation uses crowded tournament selection (preferring lower rank, then larger crowding distance), uniform crossover with probability $p_c =0.8$, and per-gene pointwise mutation with probability $p_m=0.2$ drawn from the gene’s domain.

        \item Offspring are evaluated and merged with parents; we apply fast non-dominated sorting where the candidate solutions are ranked based on Pareto dominance to balance the trade-offs between multiple objectives. This ensures that solutions closer to the Pareto front are favored.

        \item Then the truncation is applied based on crowding-distance to retain $P$ elites. The crowding distance measure preserves the diversity among non-dominated solutions and avoid premature convergence to a narrow set of architectures.

        \item We terminate after $G=8$ generations or early if the rank-$0$ front remains unchanged for $T=2$ consecutive generations.
    \end{itemize}

\section{Results and Discussion}

We evaluated the proposed FLOPs-aware HQNN architecture search on five benchmark datasets: MNIST, Wine, Breast Cancer, Iris, and Digits. In all cases, the input and output layers were fixed, which are dependent on input size and number of classes in the dataset, respectively. The neurons in the first and only hidden layer scales with the number of qubits in the PQC. This ensures that the majority of variability explored during search was due to the quantum circuits, making FLOPs-awareness a central criterion in guiding efficient architecture discovery. All candidate architectures were trained for $5$ epochs using the Adam optimizer with a learning rate of $0.003$, and a batch size of $64$ across all datasets. Figures \ref{fig:results_mnist}-\ref{fig:results_BC} present the results, where accuracy is plotted against classical, quantum, and total FLOPs. Candidate architectures are shown in purple, while Pareto-optimal points are highlighted in gold.

\subsection{Results on MNIST Dataset}

\begin{figure*}[htpb]
    \centering
    \includegraphics[width=0.99\linewidth]{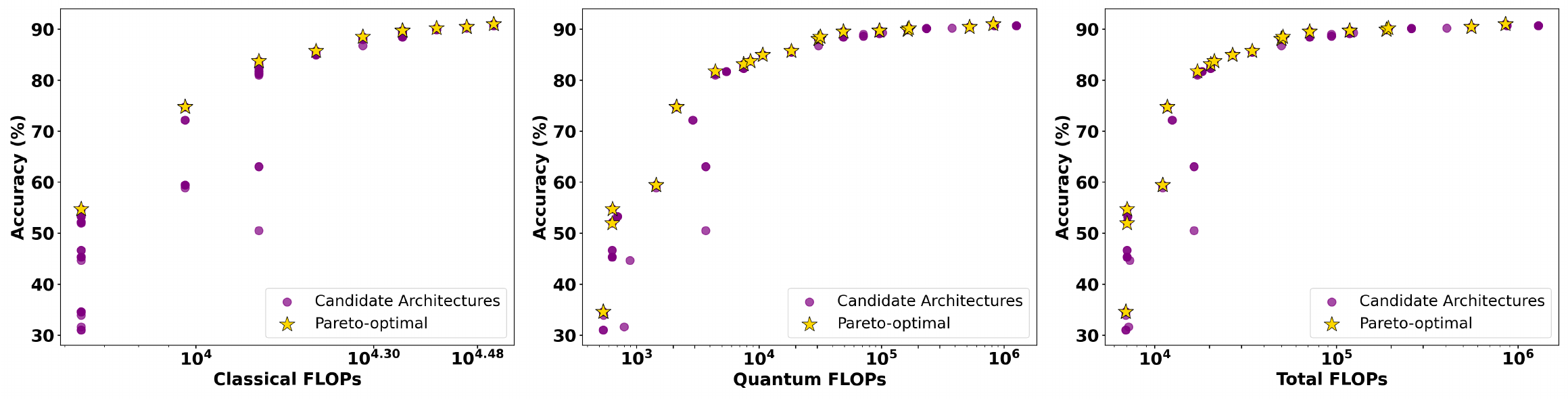}
    \vspace{-6pt}
    \caption{Accuracy versus computational cost for candidate HQNN architectures on the MNIST dataset. Each panel shows accuracy as a function of (left) classical FLOPs, (middle) quantum FLOPs, and (right) total FLOPs. Purple dots represent all candidate architectures while gold stars denote Pareto-optimal solutions.}
    \label{fig:results_mnist}
\end{figure*}

For the MNIST dataset, a clear dependence on computational cost is observed, as shown in Fig. \ref{fig:results_mnist}. Architectures with low FLOPs, particularly on the classical side, achieve accuracies below 50\%, but performance improves steadily with increasing computational budgets. 
Accuracy rises above 80\% once classical FLOPs exceed $~10^4$, with further gains obtained primarily through quantum FLOPs. The quantum FLOPs plot reveals a sharper trend, where accuracy increases significantly around $10^4$ FLOPs and plateaus near 90\% beyond $~10^5$ FLOPs. The total FLOPs plot closely follows the quantum FLOPs trend, underscoring that accuracy gains in MNIST are driven primarily by quantum circuit complexity. The Pareto-optimal points highlight that high accuracy can be obtained without resorting to the most resource-intensive solutions, confirming the benefit of FLOPs-aware optimization in balancing accuracy and efficiency.


\subsection{Results on Digits Dataset}
\begin{figure*}[htpb]
    \centering
    \includegraphics[width=0.99\linewidth]{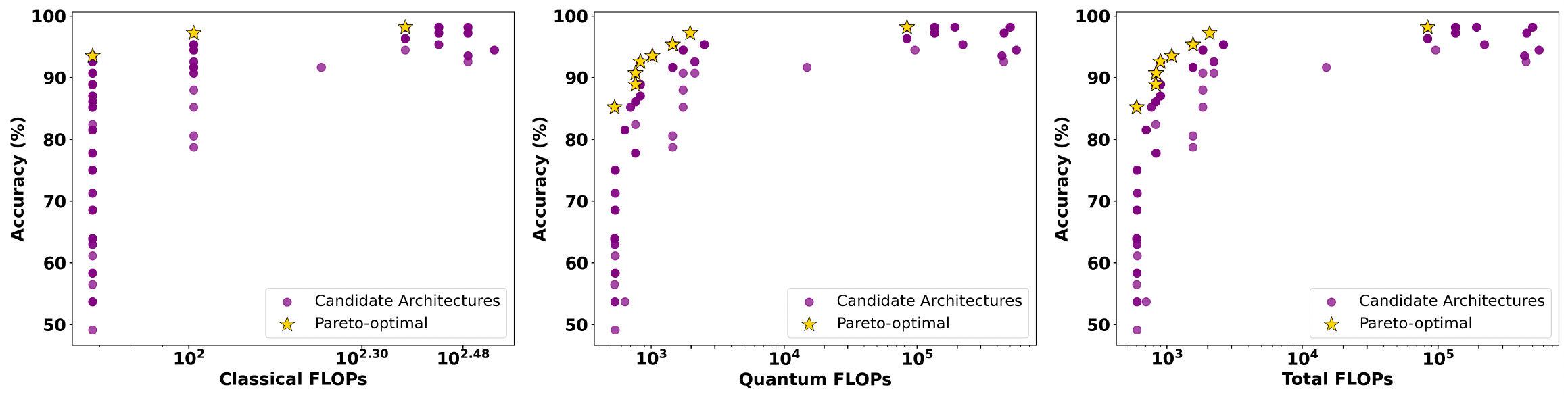}
    \vspace{-6pt}
    \caption{Accuracy versus computational cost for candidate HQNN architectures on the Digits dataset. Each panel shows accuracy as a function of (left) classical FLOPs, (middle) quantum FLOPs, and (right) total FLOPs. Purple dots represent all candidate architectures while gold stars denote Pareto-optimal solutions.}
    \label{fig:results_digits}
\end{figure*}

The Digits dataset presents a distinct behavior compared to the MNIST, as shown in Fig. \ref{fig:results_digits}. Accuracy improves rapidly with small increases in quantum FLOPs, surpassing 90\% at budgets of $~10^3–10^4$. Beyond this point, performance quickly saturates near 100\%, and larger circuits provide no additional benefit. The classical FLOPs distribution shows Pareto-optimal points at both low and high FLOPs, but with no strong dependency between accuracy and cost. The total FLOPs plot is sharply defined, with a clear Pareto front demonstrating that many architectures achieve maximal accuracy at substantially lower FLOPs. These results indicate that for moderately complex datasets like Digits, FLOPs-aware search is highly effective at identifying compact and efficient HQNN architectures without compromising performance.

\subsection{Results on Iris Dataset}
\begin{figure*}[htpb]
    \centering
    \includegraphics[width=0.99\linewidth]{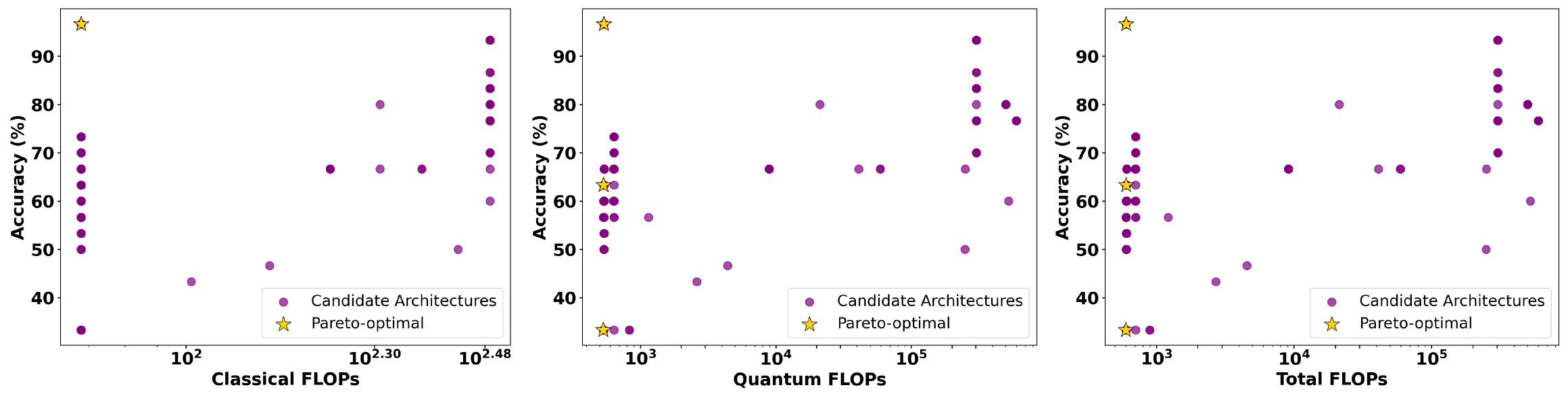}
    \vspace{-6pt}
    \caption{Accuracy versus computational cost for candidate HQNN architectures on the Iris dataset. Each panel shows accuracy as a function of (left) classical FLOPs, (middle) quantum FLOPs, and (right) total FLOPs. Purple dots represent all candidate architectures while gold stars denote Pareto-optimal solutions.}
    \label{fig:results_iris}
\end{figure*}

For the Iris dataset, accuracy outcomes vary widely, ranging from ~40\% to above 90\% across candidate architectures, as presented in Fig.~\ref{fig:results_iris}. Interestingly, the classical FLOPs plot shows that several Pareto-optimal architectures achieving high accuracy are found at very low computational budgets, suggesting that classical overhead is not a limiting factor for this dataset. The quantum FLOPs results reveal that strong performance can be achieved at both low and high FLOPs, although many mid-range FLOPs architectures perform inconsistently. The total FLOPs distribution reinforces the importance of careful search: FLOPs-aware optimization identifies efficient solutions that avoid wasted resources while still delivering high accuracy. This variability illustrates that, for simple datasets such as Iris, careful selection of quantum circuits plays a disproportionately important role in determining performance.


\subsection{Results on Wine Dataset}
\begin{figure*}[htpb]
    \centering
    \includegraphics[width=0.99\linewidth]{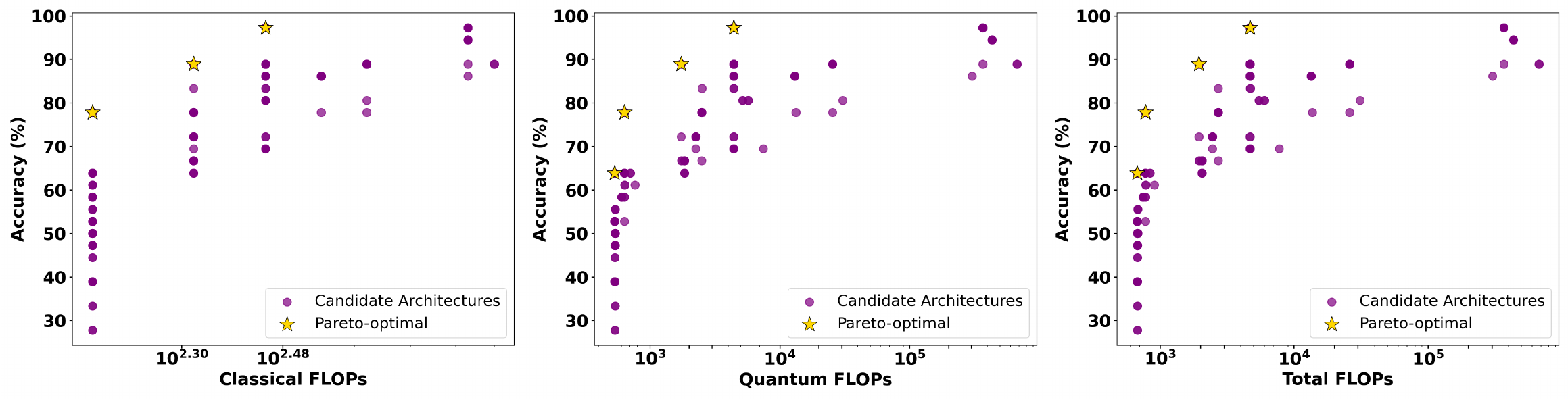}
    \vspace{-6pt}
    \caption{Accuracy versus computational cost for candidate HQNN architectures on the Wine dataset. Each panel shows accuracy as a function of (left) classical FLOPs, (middle) quantum FLOPs, and (right) total FLOPs. Purple dots represent all candidate architectures while gold stars denote Pareto-optimal solutions.}
    \label{fig:results_wine}
\end{figure*}

On the Wine dataset, the trade-off between FLOPs and accuracy is not very straightforward (Fig. \ref{fig:results_wine}). Candidate architectures span a broad accuracy range from 30\% to nearly 100\%, and high-accuracy models emerge even at modest computational budgets. The classical FLOPs plot shows that efficient solutions exist with relatively low costs, while the quantum FLOPs plot reveals a smoother trend where accuracy improves gradually but not monotonically with increased FLOPs. The total FLOPs distribution confirms that excessive computational budgets are not strictly necessary, as several Pareto-optimal solutions achieve $>$90\% accuracy while incurring far fewer FLOPs than the most complex architectures. These results indicate that for smaller datasets like Wine, FLOPs-aware search is critical for uncovering efficient solutions that avoid over-provisioned quantum circuits.


\subsection{Results on Breast Cancer Dataset}
\begin{figure*}[htpb]
    \centering
    \includegraphics[width=0.99\linewidth]{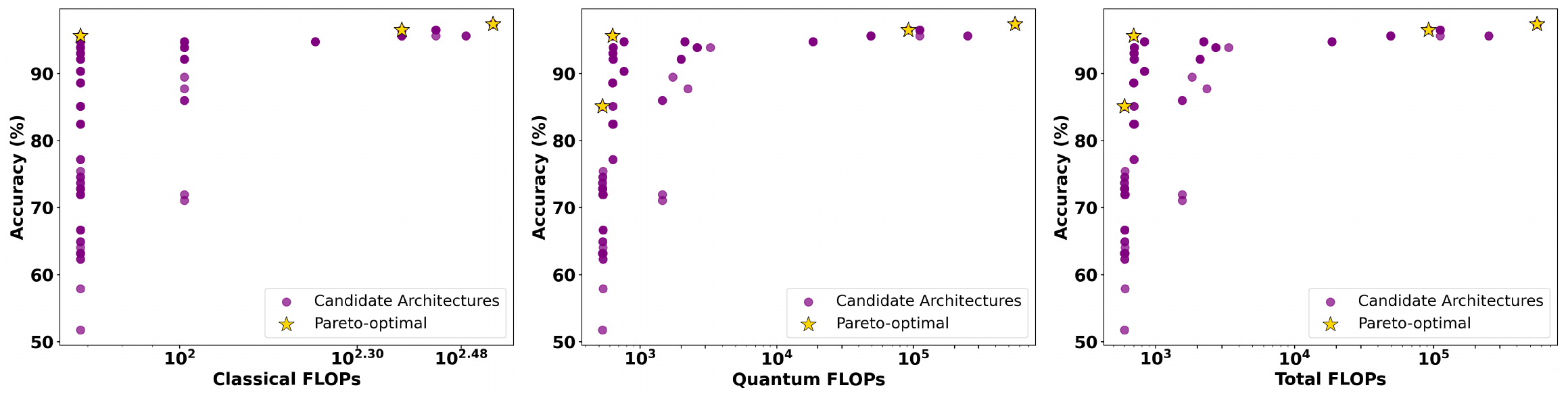}
    \vspace{-6pt}
    \caption{Accuracy versus computational cost for candidate HQNN architectures on the Breast Cancer dataset. Each panel shows accuracy as a function of (left) classical FLOPs, (middle) quantum FLOPs, and (right) total FLOPs. Purple dots represent all candidate architectures; gold stars denote Pareto-optimal solutions.}
    \label{fig:results_BC}
\end{figure*}

The Breast Cancer dataset exhibits a different trend, shown in Fig.~\ref{fig:results_BC}. Many candidate architectures already achieve strong performance, with accuracies above 70\% even at low FLOPs. Both the quantum and total FLOPs plots reveal a dense cluster of solutions achieving ~90–95\% accuracy across a wide FLOPs range. Classical FLOPs remain relatively uninformative, as even minimal classical cost yields strong performance. The Pareto-optimal front shows that high accuracy can be sustained at low or moderate FLOPs, reflecting the dataset’s relatively low complexity. In this case, FLOPs-aware search demonstrates its strength in identifying minimal architectures that retain high accuracy, confirming that more complex solutions are unnecessary.

\subsection{Comparative Analysis}
A comparative analysis across datasets highlights several consistent trends. Accuracy generally increases with FLOPs but saturates after a dataset-dependent threshold. Large-scale datasets such as MNIST require higher FLOPs budgets to reach strong performance, while moderate datasets like Digits achieve near-maximum accuracy with relatively few FLOPs. Smaller and simpler datasets, including Wine, Breast Cancer, and Iris, frequently yields efficient solutions that achieve high accuracy at surprisingly low FLOPs, though with more variability in performance among candidate architectures. In all cases, improvements in accuracy are primarily driven by quantum FLOPs, since classical FLOPs remain largely fixed due to the constrained design of classical layers.
Overall, these results demonstrate that FLOPs-aware HQNN search consistently identifies architectures that balance accuracy and computational efficiency. The Pareto fronts across datasets confirm that strong performance can be obtained without over-provisioning FLOPs, with dataset complexity dictating the threshold at which additional FLOPs cease to provide accuracy gains. The findings underscore the importance of incorporating FLOPs-awareness into HQNN search, as it not only enables scalable optimization for large datasets but also prevents wasted computational resources for simpler ones.
\section{Conclusion}
We presented a FLOPs-aware NAS framework for HQNNs, addressing the challenge of designing models that are both accurate and computationally efficient. By explicitly considering FLOPs alongside accuracy, our approach uncovers Pareto-optimal architectures that balance performance with resource cost. Results across five datasets highlight that accuracy improvements are primarily driven by quantum FLOPs, while classical FLOPs remain fixed and less influential in a hybrid setting. Importantly, FLOPs-aware analysis identifies dataset-specific thresholds beyond which additional FLOPs yield no further accuracy gains, preventing over-provisioned designs. This reveals a key intuition about HQNNs: their efficiency depends not on maximizing circuit size, but on identifying the smallest quantum circuit that reaches the accuracy plateau for a given problem. These findings underscore that FLOPs-aware design is essential for NISQ-era HQNNs, providing immediate benefits for simulator-based exploration and forward-looking value for scalable quantum–classical computing.

\section*{Acknowledgment}
This work was supported in part by the NYUAD Center for Quantum and Topological Systems (CQTS), funded by Tamkeen under the NYUAD Research Institute grant CG008.

\end{spacing}
\bibliographystyle{IEEEtran}

\bibliography{refs}

\end{document}